\documentclass[journal]{IEEEtran}
\usepackage{times}
\usepackage{epsfig}
\usepackage{float}
\usepackage{graphicx}
\usepackage{amsmath}
\usepackage{amssymb}
\usepackage{hyperref}
\graphicspath{{fig/}}

\usepackage{cite}

\ifCLASSINFOpdf
\else
\fi
\begin{document}
%
\title{MOGAN: Morphologic-structure-aware Generative Learning from a Single Image}
%
%
%

\author{Jinshu~Chen, Qihui~Xu, Qi~Kang,~\IEEEmembership{Senior~Member,~IEEE}, and~MengChu~Zhou,~\IEEEmembership{Fellow,~IEEE}
\thanks{M. Shell was with the Department
of Electrical and Computer Engineering, Georgia Institute of Technology, Atlanta,
GA, 30332 USA e-mail: (see http://www.michaelshell.org/contact.html).}
\thanks{J. Doe and J. Doe are with Anonymous University.}
\thanks{Manuscript received April 19, 2005; revised August 26, 2015.}}

%
%

\markboth{\tiny This work has been submitted to the IEEE for possible publication.Copyright may be transferred without notice, after which this version may no longer be accessible.}%
{Shell \MakeLowercase{\textit{et al.}}: Bare Demo of IEEEtran.cls for IEEE Journals}
%




\twocolumn[{%
\renewcommand\twocolumn[1][]{#1}%
\maketitle

\begin{figure}[H]
\hsize=\textwidth
\begin{center}
\includegraphics[height=8cm, width=17cm]{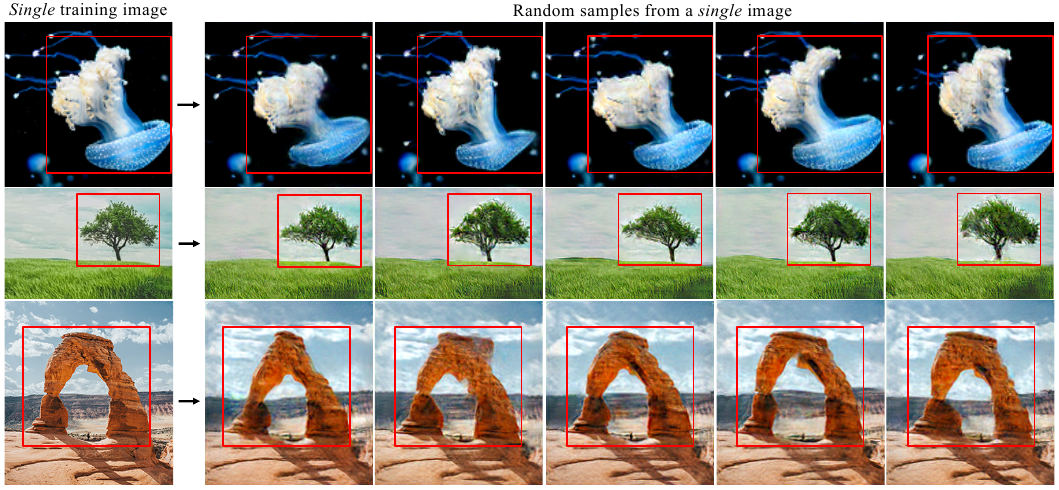}
\end{center}
	\caption{Random generation learned from a single image. We introduce an unconditional generative model which is competent for interactive ROI-based image generation tasks while based on one image only. It can generate various samples of high quality which own both rational structures and diverse appearances.}
\label{fig1}
\end{figure}
}]

\begin{abstract}
In most interactive image generation tasks, given regions of interest (ROI) by users, the generated results are expected to have adequate diversities in appearance while maintaining correct and reasonable structures in original images. Such tasks become more challenging if only limited data is available. Recently proposed generative models complete training based on only one image. They pay much attention to the monolithic feature of the sample while ignoring the actual semantic information of different objects inside the sample. As a result, for ROI-based generation tasks, they may produce inappropriate samples with excessive randomicity and without maintaining the related objects' correct structures. To address this issue, this work introduces a MOrphologic-structure-aware Generative Adversarial Network named \textup{MOGAN} that produces random samples with diverse appearances and reliable structures based on only one image. For training for ROI, we propose to utilize the data coming from the original image being augmented and bring in a novel module to transform such augmented data into knowledge containing both structures and appearances, thus enhancing the model's comprehension of the sample. To learn the rest areas other than ROI, we employ binary masks to ensure the generation isolated from ROI. Finally, we set parallel and hierarchical branches of the mentioned learning process. Compared with other single image GAN schemes, our approach focuses on internal features including the maintenance of rational structures and variation on appearance. Experiments confirm a better capacity of our model on ROI-based image generation tasks than its competitive peers.
\end{abstract}

\begin{IEEEkeywords}
Generative adversarial networks, single sample, morphologic awareness, ROI-based image generation tasks
\end{IEEEkeywords}

%
\IEEEpeerreviewmaketitle

\section{Introduction}
%
%
%
%
\IEEEPARstart{I}{n} many interactive image generation tasks, users tend to be more interested in certain targets or objects in a given sample (i.e., regions of interest or ROI) while paying less attention to the rest areas (called background). As a kind of unsupervised model, Generative Adversarial Networks (GANs)~\cite{1_Goodfellow2014GenerativeAN} are capable of most generation tasks, which have greatly promoted the development of many fields such as image inpainting~\cite{36_zheng2019pluralistic, 47_zhao2020uctgan, 48_yi2020contextual}, image-to-image translation~\cite{37_anokhin2020high, 38_zhang2020cross, 45_bhattacharjee2020dunit, 67_Isola2017ImagetoImageTW} and image synthesis~\cite{39_shocher2020semantic, 40_choi2020stargan, 46_lee2020maskgan}. However, owing to their frail structures that are hard to converge, GANs heavily depend on large datasets or plenty of prior knowledge to complete their training, thus making it an obstacle for GANs to get widely utilized.

In many cases, it is hard to get access to sufficient data of high quality to meet a GAN's training needs. Under such circumstances, effectively learning robust features from a few samples (or a single sample at an extreme case~\cite{2_Zontak2011InternalSO}) has become a crucial challenge. Besides many classical tasks~\cite{3_Zontak2013SeparatingSF, 4_Michaeli2014BlindDU, 5_Bahat2016BlindDU, 6_Freedman2011ImageAV}, recently proposed models~\cite{7_Shaham2019SinGANLA, 8_hinz2021improved, 9_Gur2020HierarchicalPV} can accomplish image generation tasks efficiently by adopting hierarchical GAN pyramid structures~\cite{18_karras2019style, 41_karras2018progressive, 42_wang2018high, 43_denton2015deep, 44_zhang2017stackgan}, making it possible for unconditional GANs to generate various samples based on only one image. Nevertheless, such models treat different patches of an image equally regardless of their actual semantic information. Given ROI, the models mentioned tend to generate confusing results because they neither provide any interface for users to specify objects as ROI or background nor can they guarantee that the specific objects own proper structures in their generated results. Intuitively, these models produce multiform fake samples by choosing some random patches and applying ``copy-shift-paste'' operations, which are quite likely to destroy the rational structures of objects inside the sample. 

To overcome such deficiency, we follow the same basic precondition (i.e., rely on only one natural image) and propose a novel MOrphologic-structure-aware GAN named \textbf{MOGAN}. Our target is to acquire samples that are abundant in appearance diversities while keeping the original structures of objects inside the sample correct. Besides the image, our model takes sets of coordinates specified by users as input to distinguish between ROI and background. Inside the model, we set up two parallel branches to generate ROI and background separately, which are both organised in a hierarchical way but own different characteristics. For the ROI branch, under the premise that no extra prior data is introduced, we propose a method that augment the original image into different forms and such augmented data can be used to learn ample knowledge of both correct structures and morphological patterns. We design a lightweight style extraction module to learn an affine transform from such data then act on the original dataflow, thus providing a guidance for the generation process. This module can be trained end-to-end along with the whole model. For the background branch, the sample for learning is the rest of the image excluding the given ROI. We apply a binary mask to the original image to shield pixels in the position of ROI and thereby switch the task of generating a complete image to generating an image with a mask, which reduces the difficulty of generation to a certain extent. 

Finally, we analyze MOGAN's capabilities of managing different tasks including ROI-based random image generation, image editing and single image animation. We test and compare MOGAN with other models in terms of generated results' quality. By analyzing results qualitatively and quantitatively, this work shows that our proposed model achieves better performance than its peers. Moreover, we investigate the effects of different components on our method's performance by conducting several ablation experiments.

In summary, this work aims to make two contributions:

\textbf{1.}	On the condition of a single image for training, we propose a novel method for generative models to gain morphologic diversities while maintaining correct structures. It utilizes the morphologic information coming from the augmented original image and we design a lightweight style injector to inject such knowledge to the model.

\textbf{2.}	To well accomplish ROI-based image generation tasks (generating images according to regions users are interested in), we introduce a novel model with parallel branches to handle the concurrent and separate generation of ROI and background. In this way we manage to generate various realistic images with only one image to learn from.

In addition, this work analyzes MOGAN's abilities of managing different interactive image generation tasks such as generating random samples based on a single image, image editing and animation. Experiments are performed to validate that MOGAN achieves better performance on the quality of generated samples than its peers.

\begin{figure*}
\centering
\includegraphics[height=10.6cm, width=18cm]{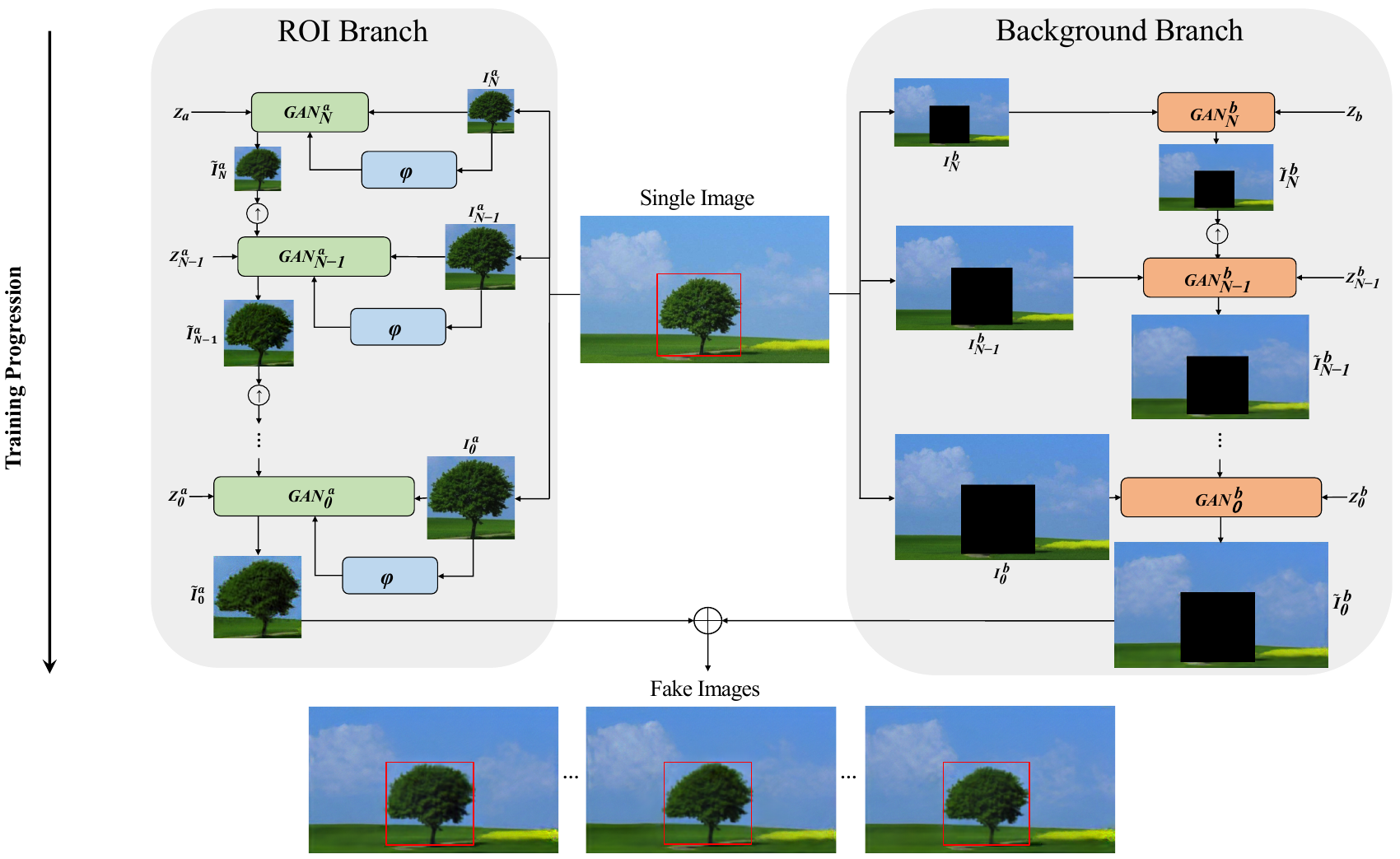}
	\caption{MOGAN contains two parallel hierarchical branches responsible for the generation of ROI and background. The ROI branch takes ROI cut from the original image as the training target while the background branch takes the original image with a binary mask standing for regions of background. Finally, the generated results produced from two branches can be fused into complete images which are of high quality.}
	\label{fig2}
\end{figure*}

\section{Related Work}

\subsection{GANs for Image Processing}

Deep learning methods have excellent performances in image generation and feature extraction. Goodfellow et al. proposed a Generative Adversarial Network (GAN)~\cite{1_Goodfellow2014GenerativeAN} based on the idea of a zero-sum game. GAN has become an important branch in the area of deep learning. It has a wide range of applications in the field of image processing~\cite{10_Zhu2016GenerativeVM, 11_Dekel2018SparseSC, 12_zhu2017unpaired, 13_chen2018sketchygan, 14_wang2016generative, 15_yu2018generative, 16_perarnau2016invertible, 17_gu2020image, 49_xiang2019single}. Image generation is the most common one. For image generation tasks, the characteristic of GANs that generating based on noise makes the GANs' generated results diverse. More and more techniques and tricks have also been proposed to enhance the stability of GANs' training process~\cite{59_Lee2009ConvolutionalDB, 60_Arjovsky2017WassersteinGA, 61_White2016SamplingGN, 62_Mao2017LeastSG}.

 However, most GANs used for image generation tasks rely on specific datasets or pre-trained models~\cite{18_karras2019style, 19_brock2018large}. Getting GANs well-trained on specific datasets places restrictions on GANs' flexibility for different tasks. Instead of extracting the common features of different images, this work focuses on the information contained in a single natural image.

\subsection{Single-image GANs}

In recent years, researchers have used the information contained in a single image to build deep learning models, thereby solving the problem of inadequate training data in some cases. InGAN~\cite{20_shocher2018ingan} represents the first such model that applies GANs to the processing of a single image. However, it is a conditional generative model. Its generation process requires specific images as input (i.e., mapping the image to an image), thus leading to its poor generalization and failure to generate images randomly. To overcome its drawbacks, Tamar et al.~\cite{7_Shaham2019SinGANLA} proposed SinGAN. It realizes random generation based on a single image by using an unconditional generation model (i.e., mapping from random noise to an image). Hence, it is suitable for many different image processing tasks. Then ConSinGAN~\cite{8_hinz2021improved} was proposed with a series of improved techniques for training single-image unconditional GANs. HP-VAE-GAN~\cite{9_Gur2020HierarchicalPV} was designed for single-image video generation. But the last three models~\cite{7_Shaham2019SinGANLA, 8_hinz2021improved, 9_Gur2020HierarchicalPV} share serious defects like excessive randomness and uncontrollability on ROI-based tasks. Compared with them, our proposed model can generate random samples with correct structures and variable appearances based on a single image and users' expectations. It broadens the application range of single image generation.

\section{Method}

We first give a brief introduction to SinGAN~\cite{7_Shaham2019SinGANLA} along with a short discussion about its limitations. Then we introduce MOGAN in detail.

\subsection{SinGAN}

SinGAN is a kind of unconditional GAN (generating samples from latent vectors). It is able to generate diverse samples from randomly sampled noise based on only one single natural image. Obeying the design of a hierarchical structure, it stacks several sub-GANs into a pyramid structure. All sub-GANs share exactly the same structure but not parameters. It takes a latent vector sampled from a Gaussian distribution along with the generated result from the previous one as input (by noting that the input of the first sub-GAN is the latent vector only). The training target of each sub-GAN is the original natural image downsampled to different sizes. Such a design makes the learning goal of the overall model gradually shift from small-scale samples which are rich in global structural information to large-scale samples which contain plenty of texture details, deepening the model's comprehension of a given sample.

However, SinGAN's generative process fails to distinguish between different areas or instances inside a given image. Qualitatively, the embodiment of the generated samples' diversity seems like randomly choosing patches from the original image, copying, randomly shifting and then pasting. As a consequence, such a random ``copy-shift-paste'' type of generation is quite likely to break down an object's structure, thereby resulting in possible irrational outcomes. In many interactive generation tasks, there are some regions in an image in which users are more interested (i.e., ROI). It is thus required for a method to generate random samples with as many changes as possible but no changes in the original semantic structures. For example, if a tree is ROI that interest users, the shape of its crown or bending angle of its trunk may change, while neither the crown nor the trunk should disappear, and the semantic structure that ``the crown is above the trunk'' should not change either. Such ROI-based tasks are hard for SinGAN to complete.

\begin{figure*}
\begin{center}
\includegraphics[height=7cm, width=18cm]{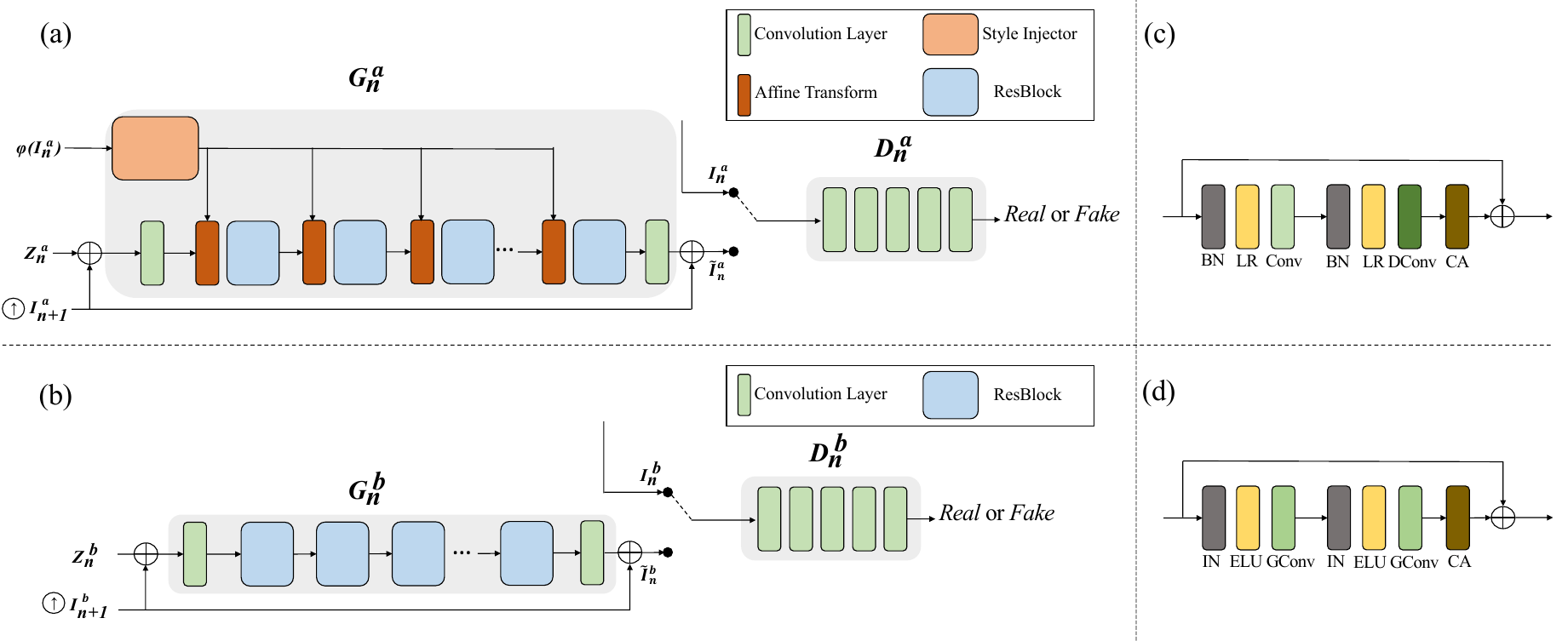}
\end{center}
	\caption{Details of sub-GANs in two branches. (a) ROI branch. Generators are organised based on residual blocks mainly contain a convolution layer and a deformable convolution layer. A novel module named a style injector that transforms the augmented original image into knowledge of structures and appearances controls the style of generation through affine transforms. (b) Background branch. Generators are built based on residual blocks mainly containing two gated convolution layers. Discriminators of both branches are Markovian discriminators. Details of residual blocks in ROI branch and background branch are shown in (c) and (d) respectively.}
\label{fig3}
\end{figure*}

\subsection{Proposed MOGAN}
\label{sec3.2}
Motivated by the problems mentioned above, we introduce our MOGAN with its structure shown in Fig.~\ref{fig2}.

\textbf{Parallel-branch architecture:} In our problem setting, the usable information includes a single natural image \textit{I} and a series of coordinates $ \{(x_1^{min}, y_1^{min}, x_1^{max}, y_1^{max}), $ ...,$ (x_m^{min}, y_m^{min}, x_m^{max}, y_m^{max})\} $ provided by users to mark \textit{m} regions to which they pay attention. Allowing the model to deal with ROI and background areas separately leads to the problem of disentanglement. Under the circumstance that the number of learnable samples is limited to one, many existing methods of disentanglement based on extra specific prior data~\cite{21_ma2018disentangled, 22_nguyen2019hologan, 23_pumarola2018unsupervised} are not applicable. Therefore, we use two different latent vectors marked as $ Z_a $ and $ Z_b $ to be responsible for the generation of ROI and background in turn similarly to~\cite{24_kwak2016generating, 25_singh2019finegan, 26_li2020mixnmatch}. Since our requirements for the generation of ROI and background are often different, the structures in charge of generating ROI and background should be independent of each other such that special adjustments can be applied to them, respectively. Therefore, we set up two parallel branches that take the two latent vectors mentioned above as input respectively and conduct their generation processes separately. In most cases where ROI can be well separated from the background, the final generated results of the two branches can be fused together directly. For some complex samples, there may be the discontinuity on the boundary. To handle it, we may take simple interpolation methods as post-processing on the edges. Note that our parallel architecture can handle the two parts' generation synchronously while the methods in~\cite{25_singh2019finegan, 26_li2020mixnmatch} have to deal with a generation process in turn. Compared with~\cite{56_Wu2020CascadeEP}, our MOGAN learns the style during the training while ~\cite{56_Wu2020CascadeEP} uses pre-extracted style expression. Moreover,~\cite{56_Wu2020CascadeEP} regards style information as explicit targets to supervise the training while we only treat the style information as guidance.

\begin{figure}[h]
\begin{center}
\includegraphics[height=4cm, width=8cm]{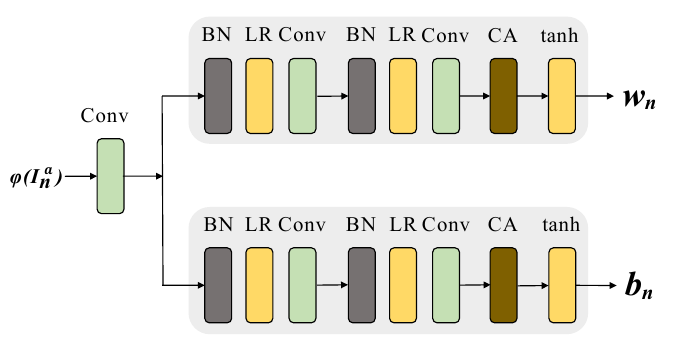}
\end{center}
	\caption{Details of a style injector. It is a lightweight encoder essentially, which contains two bypasses producing a weight and a bias respectively. Taking the augmented original image as the input, it controls the changing direction of the generator’s dataflow through affine transforms.}
\label{fig4}
\end{figure}

\textbf{ROI Branch:} For this branch, we expect the generated results to own sufficient morphological changes under the premise that the semantic structure remains correct. The learning target of the ROI branch is the ROI part cut from \textit{I} according to the coordinates mentioned above, which is marked as $ I_a$. The overall framework of the branch obeys a hierarchical design similar to SinGAN's~\cite{7_Shaham2019SinGANLA}. To be more specific, we organise a number of sub-GANs $ \{\mathit{GAN}_0^a,$...,$\mathit{GAN}_N^a\} $ and stack them into a pyramid. $\mathit{GAN}_n^a$ consists of a generator $G_n^a$ and a discriminator $D_n^a$ which are trained adversarially. Meanwhile, we set up an image pyramid $\{I_0^a,$...,$I_N^a\}$ by downsampling $I_a$ for \textit{N} times based on the rescale factor $r^N$, where $r\textgreater1$. The training process starts from the smallest scale of the image pyramid, i.e., $I_N^a$ along with $\mathit{GAN}_N^a$, and the level of training scales goes up when the previous scale has finished training. For each $I_n^a$ with the corresponding $\mathit{GAN}_N^a$, $G_n^a$ learns to map the sum of a noise $Z_n^a$ randomly sampled from a Gaussian distribution and the upsampled output of $G_{n+1}^a$ into a fake sample. To stabilize the training process, we add the output of the generator and the upsampled output of $G_{n+1}^a$ together as the final generated result marked as $\tilde{I}_n^a$. $D_n^a$ attempts to tell $I_n^a$ and $\tilde{I}_n^a$ apart. Note that the input to $G_N^a$ is $Z_a$ only and each $\mathit{GAN}_n^a$ is frozen after being trained. 

Here comes the problem: how to increase the diversity of generated results without destroying the semantic structure of the ROI. Without introducing additional training data or prior knowledge such as pre-trained models, we notice that there contains plenty of diverse morphologic information along with rational semantic structures inside the augmented original images. As a guidance, such augmented data not only shows feasible changing directions about appearance but also emphasizes the structure information for the generator. Thus, during the training of each $\mathit{GAN}_n^a$, we transform $I_n^a$ into different forms through regular data-augmentation methods including flipping vertically and horizontally, padding and randomly rescaling and so on, which are marked as $\varphi$ in a general way. In order to extract useful information from $\varphi(I_n^a)$ , we build an extra lightweight module named a style injector which takes $\varphi(I_n^a)$ as input and outputs a weight $w_n$ and a bias $b_n$. Marking the style injector for $\mathit{GAN}_n^a$ as $SI_n^a$, the process mentioned can be described as:

\begin{equation}
[w_n, b_n]=SI_n^a(\varphi(I_n^a))
\end{equation}

Next, we apply the learned affine transform on $G_n^a$'s original dataflow, guiding the generator towards an expected generation direction provided by $\varphi(I_n^a)$. The learned $w_n$ will be multiplied on the $G_n^a$'s original dataflow and the learned $b_n$ will be added. As for the structure of the module, the two bypasses responsible for $w_n$ and $b_n$ have the same formation based on residual design but do not share parameters. The architecture of the module is shown in Fig.~\ref{fig4}. Note that our module can be trained end-to-end together with the whole model. This is different from the methods in~\cite{18_karras2019style, 27_huang2017arbitrary, 28_karras2020analyzing} that all need pre-training. The total generation process can be expressed as:

\begin{equation}
\tilde{I_n^a} = \begin{cases}
G_n^a(Z_a, \varphi(I_n^a)), & n=N \\
G_n^a(Z_n^a, \varphi(I_n^a), (\tilde{I}_{n+1}^a\uparrow^{upsample})), & n<N \\
\end{cases}
\end{equation}

\begin{figure*}
\begin{center}
\includegraphics[height=22cm, width=17cm]{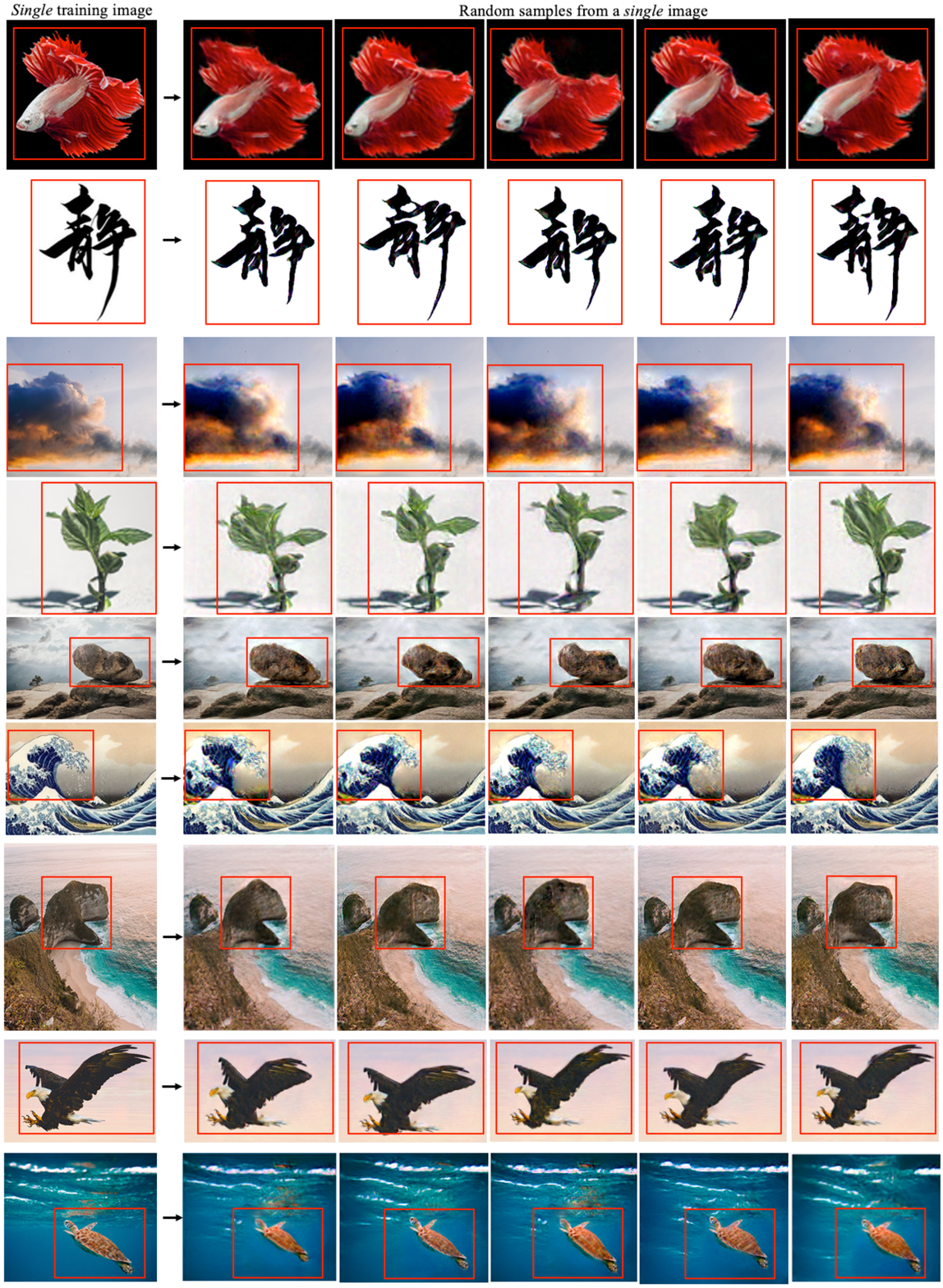}
\end{center}
	\caption{Randomly generated samples. Our model can produce diverse images across different areas and topics for ROI-based image generation tasks. The generated results maintain the original structure of the objects while get plenty of changes on appearance.}
\label{fig5}
\end{figure*}

\begin{figure*}
\begin{center}
\includegraphics[height=7cm, width=18cm]{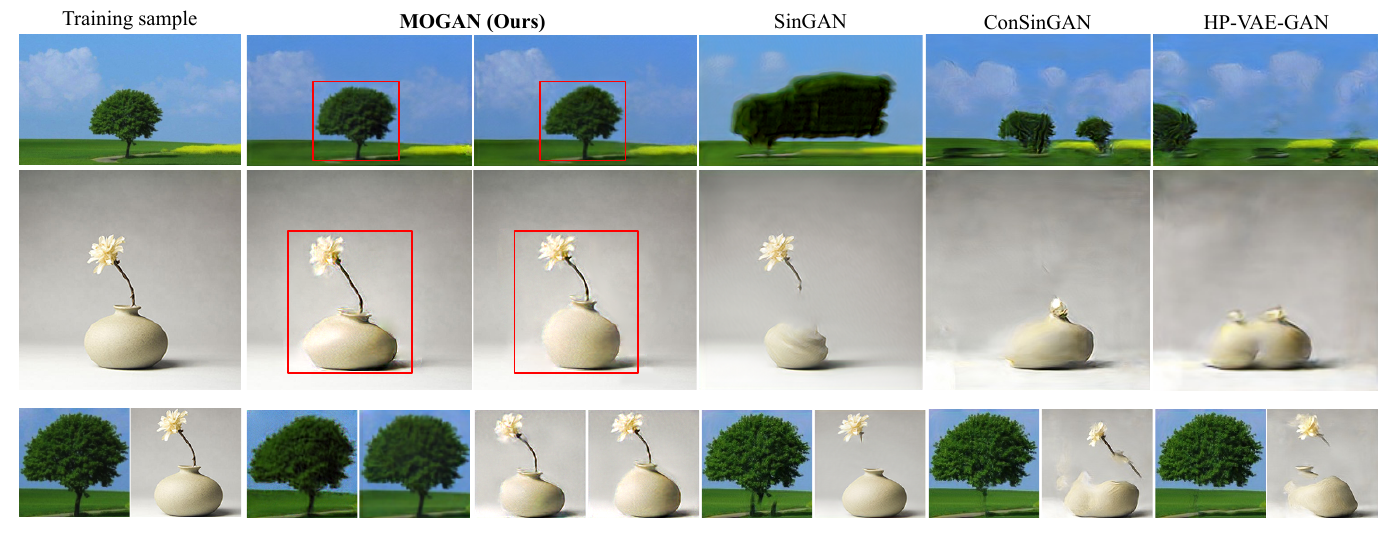}
\end{center}
	\caption{Randomly generated samples from SinGAN, ConSinGAN, HP-VAE-GAN and ours. For the whole image, other models tend to generate blurry or confusing results. For the ROI part only, other results easily get stuck in overfitting or irrational outcomes while ours are changeable and stable.}
\label{fig6}
\end{figure*}

\begin{table*}[t]
\centering
	\caption{Scores of average SIFID and average diversity calculated on samples generated by SinGAN~\cite{7_Shaham2019SinGANLA}, ConSinGAN~\cite{8_hinz2021improved}, HP-VAE-GAN~\cite{9_Gur2020HierarchicalPV} and MOGAN. Results show that our model achieve the best performance than others.}
\label{tab1}
\begin{tabular}{ccccc}
	\hline Metrics & SinGAN~\cite{7_Shaham2019SinGANLA} & ConSinGAN~\cite{8_hinz2021improved} & HP-VAE-GAN~\cite{9_Gur2020HierarchicalPV} & \textbf{MOGAN (Ours)}\\
	\hline SIFID (whole) & 0.72 & 0.63 & 0.61 & \textbf{0.22}\\
	Diversity (whole) & 0.42 & 0.49 & \textbf{0.51} & 0.20\\
	GQI (whole) & 0.58 & 0.78 & 0.84 & \textbf{0.91}\\
	SIFID (ROI-only) & 0.19 & 0.59 & 0.56 & \textbf{0.11}\\
	Diversity (ROI-only) & 0.21 & \textbf{0.51} & 0.50 & 0.39\\
	GQI (ROI-only) & 1.11 & 0.86 & 0.89 & \textbf{3.55}\\
	\hline
\end{tabular}
\end{table*}

Now that we have brought in more information which may not only lead to better training results but confuse the generator as well. We then consider a more robust design for $G_n^a$ . In order to handle all sizes of samples, we build $G_n^a$ based on a fully convolutional architecture. We set convolution layers at the beginning and end of $G_n^a$, while we place several residual blocks~\cite{63_He2016DeepRL} in the middle. Since interference caused by $\varphi(I_n^a)$ is hard to eliminate, we then treat it as a strong noise. Taking the vacant areas appearing at the four corners of the image after rotating for example, such interference can severely disturb the data distribution in certain specific areas, which is harmful and different from randomly sampled noise $Z_n^a$ that uniformly acts on the dataflow. To cope with this issue, we change the last convolution layer inside each residual block to a deformable convolution layer~\cite{29_dai2017deformable, 30_zhu2019deformable} and add a channel-wise attention layer~\cite{31_qilong2020eca} behind it, which makes $G_n^a$ focus on valuable regions and overlook other disturbing information. Before each convolution layer and deformable convolution layer, we set a BatchNorm layer~\cite{64_Ioffe2015BatchNA} and a LeakyReLU layer. Another advantage of our design is that we can easily arrange where a style injector plays a part. To be specific, we set a style injector right in front of every residual block, thus ensuring that both the learnable details and bad influence can be handled in a timely fashion. Details of $\mathit{GAN}_n^a$ are shown in Fig.~\ref{fig3}a and Fig.~\ref{fig3}c.

As for other structure and training details of the ROI branch, $D_n^a$ is a Markovian discriminator~\cite{32_li2016precomputed, 33_isola2017image} with a fully convolution structure. For the discriminators, besides the traditional adversarial loss marked as $L_0$, we use WGAN-GP loss~\cite{34_gulrajani2017improved} marked as $L_{WGAN-GP}$ to stabilize $D_n^a$'s training process. WGAN-GP loss can be expressed as:

\begin{equation}
L_{WGAN-GP}=(\left\|\nabla_{\tilde{I_n^a}}D(\tilde{I_n^a})\right\|-1)^2
\end{equation}

The final loss function for the discriminators can be expressed as:

\begin{equation}
L_D=L_0(G_n^a,D_n^a)+\lambda{L_{WGAN-GP}}(D_n^a)
\end{equation}

For $G_n^a$, besides traditional adversarial loss, we choose mean squared error (MSE) and cosine distance together as loss function. They measure the similarity between $\tilde{I}_n^a$ and $I_n^a$ in different aspects: cosine distance marked as $L_1$ emphasizes the coherence of global direction and may tolerate local structure difference to some extent. It can be described as:

\begin{equation}
L_1=1-cos(\tilde{I_n^a}, I_n^a)
\end{equation}

 MSE marked as $L_2$ requires pixel-level consistency to restrain $G_n^a$ from generating bad texture futures. It can be described as:

\begin{equation}
L_2=\left\|\tilde{I_n^a}-I_n^a\right\|^2
\end{equation}

The final loss function for the generators can be expressed as:

\begin{equation}
L_G=L_0(G_n^a,D_n^a)+\alpha{L_1}(G_n^a)+\beta{L_2}(G_n^a)
\end{equation}

\textbf{Background Branch:} Actually, background regions of different samples may vary considerably in complexity. In other words, some samples may own extremely simple background even pure color while others may contain various disparate instances. Since users are less interested in the background, we expect background generation to maintain the global uniformity primarily while making changes in a few local patches, thus no need to introduce extra modules like a style injector into background generation. It doesn't mean it's unnecessary for the generated background to change. For some tasks like image-manipulating task, we believe that only changes in ROI are needed. However, ROI-based tasks also include certain tasks like the data augmentation task where changes of the background are expected: Samples with diversity in both ROI and background are more beneficial for the following CV tasks than those with diversity in only the ROI part. We aim to make our method more general for ROI-based tasks and we believe our "two-branch" idea gives a more proper way.

To isolate the influence of ROI, we apply a binary mask to \textit{I} and mark the result as $I_b$. Similar to methods of ROI, we organise another GAN pyramid $ \{\mathit{GAN}_0^b,$...,$\mathit{GAN}_N^b\} $ along with an image pyramid $\{I_0^b,$...,$I_N^b\}$, where $I_n^b$ becomes the training target of $\mathit{GAN}_n^b$. Then the generation target can be approximated to a less-difficult image inpainting task for areas inside the mask which are eventually replaced by the generated results of the ROI branch. Similar to the ROI branch, the generation process can be described as:

\begin{equation}
\tilde{I_n^b} = \begin{cases}
G_n^b(Z_b), & n=N \\
G_n^b(Z_n^b, (\tilde{I}_{n+1}^b\uparrow^{upsample})), & n<N \\
\end{cases}
\end{equation}

For the generator in $\mathit{GAN}_n^b$ marked as $G_n^b$, we adopt similar residual-type of design as mentioned in the ROI's branch but replace all the convolution layers and deformable convolution layers with gated convolution layers~\cite{35_yu2019free}. This enables the model to learn the soft mask while learning the pixels outside the mask. Besides, we replace the BatchNorm-LeakyReLU layers used in the ROI branch with InstanceNorm-ELU~\cite{65_Ulyanov2016InstanceNT} layers. Other training details including loss functions and structures of discriminators are as same as the ROI branch's. Details of $\mathit{GAN}_n^b$ are shown in Fig.~\ref{fig3}b and Fig.~\ref{fig3}d.

\textbf{Training Details:} For both the ROI branch and the background branch, we use the Adam optimizer~\cite{50_Kingma2015AdamAM} with $\beta_1=0$ and $\beta_2=0.99$. The learning rate of every generator and discriminator in both branches is set to 0.0003. For the loss function of all generators in the ROI branch, $\alpha$ is set to 50 but $\beta$ varies according to different scales: For coarse scales (such as scale N and N-1), $\beta$ is set to 10; For other scales, $\beta$ is set to 5. As for generators in the background branch, $\alpha$ is set to 50 and $\beta$ is set to 10 for all scales. In both branches, $\lambda$ for all discriminators is set to 1. We stack three ResBlocks in all generators and five convolution layers in all discriminators. The augmenting methods we take include the random vertical flip, the random horizontal flip, the random rotation, the random affine transform, the random perspective transform and the random erasing~\cite{51_Zhong2020RandomED}.


\begin{figure*}[t]
\begin{center}
\includegraphics[height=4cm, width=18cm]{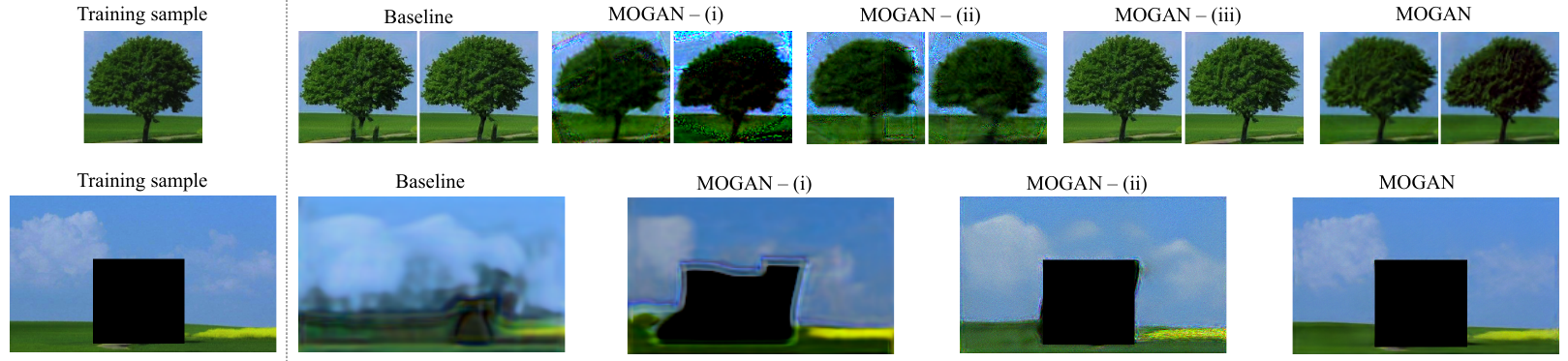}
\end{center}
	\caption{Results of an ablation study. It is clear that a style injector plays an important part in generating diversely. Deformable convolution and channel attention both make the results refiner. Gated convolution enables the model to handle data with masks.}
\label{fig7}
\end{figure*}

\begin{table*}[t]
\centering
	\caption{Quantitative results of the ablation study. Results indicate the same conclusion as Fig.~\ref{fig7}.}
\label{tab2}
\begin{tabular}{cccc}
	\hline Settings & SIFID (ROI) & Diversity (ROI) & GQI (ROI)\\
	\hline Baseline & 0.19 & 0.21 & 1.11 \\
	MOGAN - (i) & 0.21 & \textbf{0.42} & 2.00\\
	MOGAN - (ii) & 0.16 & 0.29 & 1.81 \\
	MOGAN - (iii) & 0.13 & 0.11 & 0.85 \\
	MOGAN & \textbf{0.11} & 0.39 & \textbf{3.55}\\
	\hline
\end{tabular}
\begin{tabular}{cccc}
	\hline Settings & SIFID (background) & Diversity (background) & GQI (background)\\
	\hline Baseline & 0.88 & \textbf{0.84} & 0.61 \\
	MOGAN - (i) & 0.56 & 0.63 & 1.13\\
	MOGAN - (ii) & 0.40 & 0.48 & 1.20 \\
	MOGAN & \textbf{0.24} & 0.31 & \textbf{1.29} \\
	\hline
\end{tabular}
\end{table*}

\section{Results}

We first explain MOGAN’s abilities of managing different ROI-based generation tasks along with revealing some of the results. Next, we compare the performance of our model with its peers’ qualitatively and quantitatively. Finally, we validate the effectiveness of our model’s components through an ablation study.

\subsection{Applications}

\textbf{Generating Random Samples:} \textit{To generate random samples from noise through training against a single natural image} is one of the basic capacities of our model. Samples randomly generated by our model are listed in Fig.~\ref{fig1} and Fig.~\ref{fig5} which contain diverse kinds of images that our model has enough robustness against samples of different styles and topics.

Qualitatively, the generated outcome of ROI has kept a reasonable structure by comparing it with the original image’s. In the meantime, ROI generated results get visible diversification on appearance such as shape and posture. As for backgrounds, the generated results exhibit smooth changes in local parts and retain the global layout similar to the original sample’s. Note that the model shows the prominent effects on samples that have more freedom degrees to change on appearance. As for samples with a solid structure which is hard to change, the effects seem to reveal in the aspects of postures.

It is worth noting that, we use the raw augmented data (not well pre-trained models or any other specific prior knowledge like some existing methods~\cite{57_Gu2020PriorGANRD, 58_Shocher2020SemanticPF}) to enhance structure awareness. However, the raw data contains strong noise as described in sec.~\ref{sec3.2}. Injecting such noise into the model may disturb the training and cause blur results. As a result, intuitively, we may find slight shadow or artifacts in the generated results. It is unavoidable to an extent and a kind of trade-off in our opinions. To address this issue, we think that it would help if limiting the effects of the style injecting, e.g., multiplying a factor \textit{c} ($0<c<1$) on \textit{w} and \textit{b} produced by the style injector before they are applied onto the model; setting more ResBlocks between every two affine transforms.

\begin{figure*}[t]
\begin{center}
\includegraphics[height=3.3cm, width=18cm]{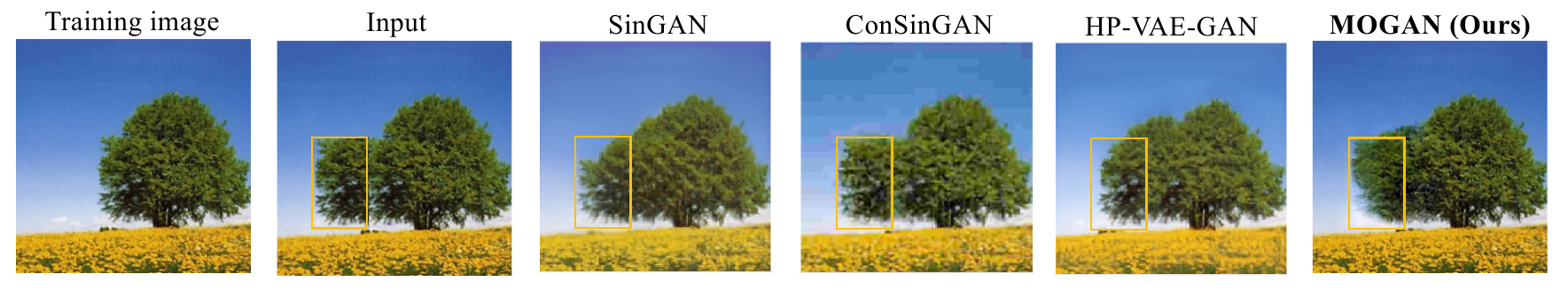}
\end{center}
	\caption{Results of the single image editing task. Boxes in yellow stand for the original position of the edited patch in the input image. Intuitively, the results of MOGAN are more vivid on the edges of the edited patch than others. The texture of MOGAN's generated results is the finest than its competitive peers'. }
\label{fig8}
\end{figure*}

\begin{figure*}[t]
\begin{center}
\includegraphics[height=2.2cm, width=18cm]{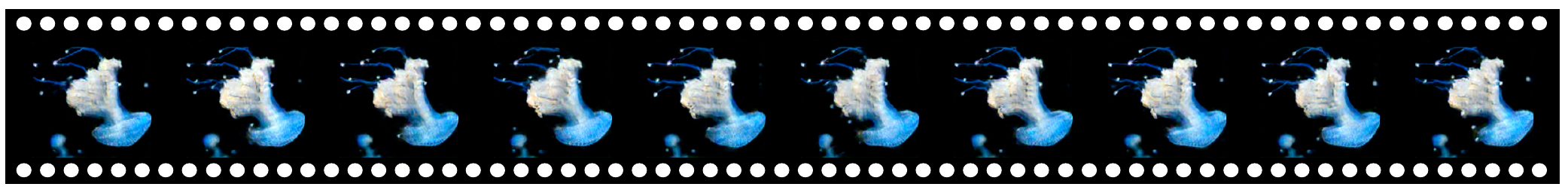}
\end{center}
	\caption{One sample result of single image animation. We obtain such samples changing smoothly by adjusting the level of augmenting gradually.}
\label{fig9}
\end{figure*}

\textbf{Editing:} \textit{To select some of the patches and paste them onto the other location of the original image, then output a harmonious result.} We notice the target of image editing tasks exactly fits the capacity of our model. To perform editing, we take the image pasted with edited patches as the input of a style injector. In this way, the edited information works similarly to the augmented data. For more details, we first train a MOGAN against the image for editing. Then we freeze all trainable parameters of the model, input the image with edited patches to the style injector and start a forward process of the model.

\textbf{Single Image Animation:} \textit{To generate a short video based on a single image,} which is an extension of the ability of random image generation. For MOGAN, after being well-trained on a certain sample, we fix the type of data-augmented methods and adjust the level of the methods gradually. For example, we enlarge the rotation angles of augmented samples by degrees during a series of image generating processes. In this way, we can obtain a number of generated results that change smoothly and organise them into the form of video. Fig.~\ref{fig9} show a sample result.

\subsection{Comparison}
We make a comparison among the state-of-the-art models based on a single image. Fig.~\ref{fig6} and Fig.~\ref{fig8} show the comparison results of randomly generating tasks and editing tasks respectively.

Qualitatively, for the randomly generating task, SinGAN and other related models have the similarity in treating samples as a whole. They tend to equally deal with all objects inside as analogical patches regardless of their different importance and semantic information, which results in blurry and ambiguous results. But our model manages the problem very well by setting two parallel branches and generating ROI and the background separately. Next, we take samples that only contain ROI as the training targets and conduct experiments again. From the results, we can find that outcomes of the other models always get stuck in two situations: overfitting (excessively similar to the training target) or meaningless (structure of objects being destroyed). The basic reason of such results is that such models lack a proper guidance for added noise. Restrained by MSE, the noise in the end either affect the texture slightly, which makes the result seem like overfitting, or affect the structure which deforms the related objects too much. On the contrary, our model preserves the original structure to a large extent while making various changes emerging on the object owing to the style injector. Qualitative experiments in our paper are mostly conducted on the Unsplash Dataset due to its high quality. Note that we have not compare the generated results of background because other models cannot deal with data with masks, which will be explained at Sec.~\ref{sec4.3}. For the editing task, intuitively, the results of ours are more vivid on the edges of the edited patch than others'. The texture of our generated results is the finest than its competitive peers'.

Quantitatively, we take Single Image Frechet Inception Distance (SIFID)~\cite{7_Shaham2019SinGANLA, 52_Heusel2017GANsTB, 66_Zhang2019SelfAttentionGA} as a metric. We take 50 different samples for models to learn from and then carry out comparison tests on 100 generated results for every sample. Moreover, we calculate the diversity following the method of the coefficient of variation (CV), which is calculated via dividing the standard deviation value by the mean of the samples. Specifically, over the 100 generated samples, we first calculate the CV of the intensity values of each pixel. We average all the CVs over all pixels as the score of the diversity of all generated results upon one training sample. Finally, considering that neither low SIFID score (perhaps caused by overfitting) nor high diversity (perhaps caused by chaos) alone can present high quality of generated samples, we define a metric calculated via dividing the diversity score by SIFID, named generation quality index (i.e., GQI). Samples used for training, inference and quantitative comparison experiments come from the ImageNet Dataset~\cite{53_Deng2009ImageNetAL} and the Unsplash Dataset which are both standard and open-source datasets. We think they are more suitable for ROI-based tasks com-paring to the Places365 Dataset~\cite{54_Zhou2018PlacesA1} and Berkeley Segmentation Dataset~\cite{55_MartinFTM01} which are claimed and used in the SinGAN paper, because each sample in both datasets has a clear topic and is easy to assign ROIs. Scores of different models are recorded in Table~\ref{tab1}. Coinciding with the qualitative analysis, our model has achieved better performance than its peers given the same samples. When trained against the whole image, the other three models get higher diversity scores than ours while their SIFID and GQI are lower due to their chaotic generated results. When trained against ROI only, SinGAN always produces overfitting results which increases its SIFID and GQI while reducing the diversity to a large extent. ConSinGAN and HP-VAE-GAN continue to generate meaningless images with high diversity but low quality. Our MOGAN's results are both diverse and realistic. Besides, GQI goes up when training on ROI only because of the lower difficulty. For our model, the diversity is lower for the whole image owing to the globally similar backgrounds.

\subsection{Ablation Study}
\label{sec4.3}

To analyze our design's impact on the generation process, we take SinGAN as the baseline and conduct ablation experiments on two branches separately. The qualitative results are described in Fig.~\ref{fig7} and the quantitative results are in Table~\ref{tab2}.

For the ROI branch, improvements we make onto a structure include (i) deformable convolution layer, (ii) channel attention layer, and (iii) style injector. The effects on the generated results after removing certain design methods can be seen in Fig.~\ref{fig7}. Note that “-” means that the method is disabled and “removing deformable convolution layer” means replacing it with a full convolution layer. Obviously, the style injector plays an important role in generating diversely as models without it induce overfitting. Deformable convolution and channel attention layers both prevent the results from strong noise and stripes to a large extent, while the former plays a more effective role.

For the background branch, we introduced (i) gated convolution layers, and (ii) channel attention layers. We take away modules or methods above in turn and record the effects on the generated results in Fig.~\ref{fig7}. Similarly, “removing gated convolution layer” means replacing it with a full convolution layer. The results suggest that without gated convolution layers, the model will treat the mask as an object of certain semantic information. Thus, gated convolution layers enable the model to handle masks which isolates background from ROI. Channel attention layers make the training more stable.

\section{Conclusion}
We have introduced MOGAN, an unconditional generative model to generate random samples based on only one natural image. The generation results from our model can maintain correct structures and exhibit plenty of diversity in appearance, which is the main improvements over other recent models. As demonstrated by our experiments, MOGAN can produce samples of high quality over different kinds of images. Our future work intends to handle more detailed information of a single sample including color and texture, and to guide the generated results of ROI and background to cohere with each other.

\ifCLASSOPTIONcaptionsoff
  \newpage
\fi

\begin{IEEEbiography}[{\includegraphics[width=1in,height=1.25in,clip,keepaspectratio]{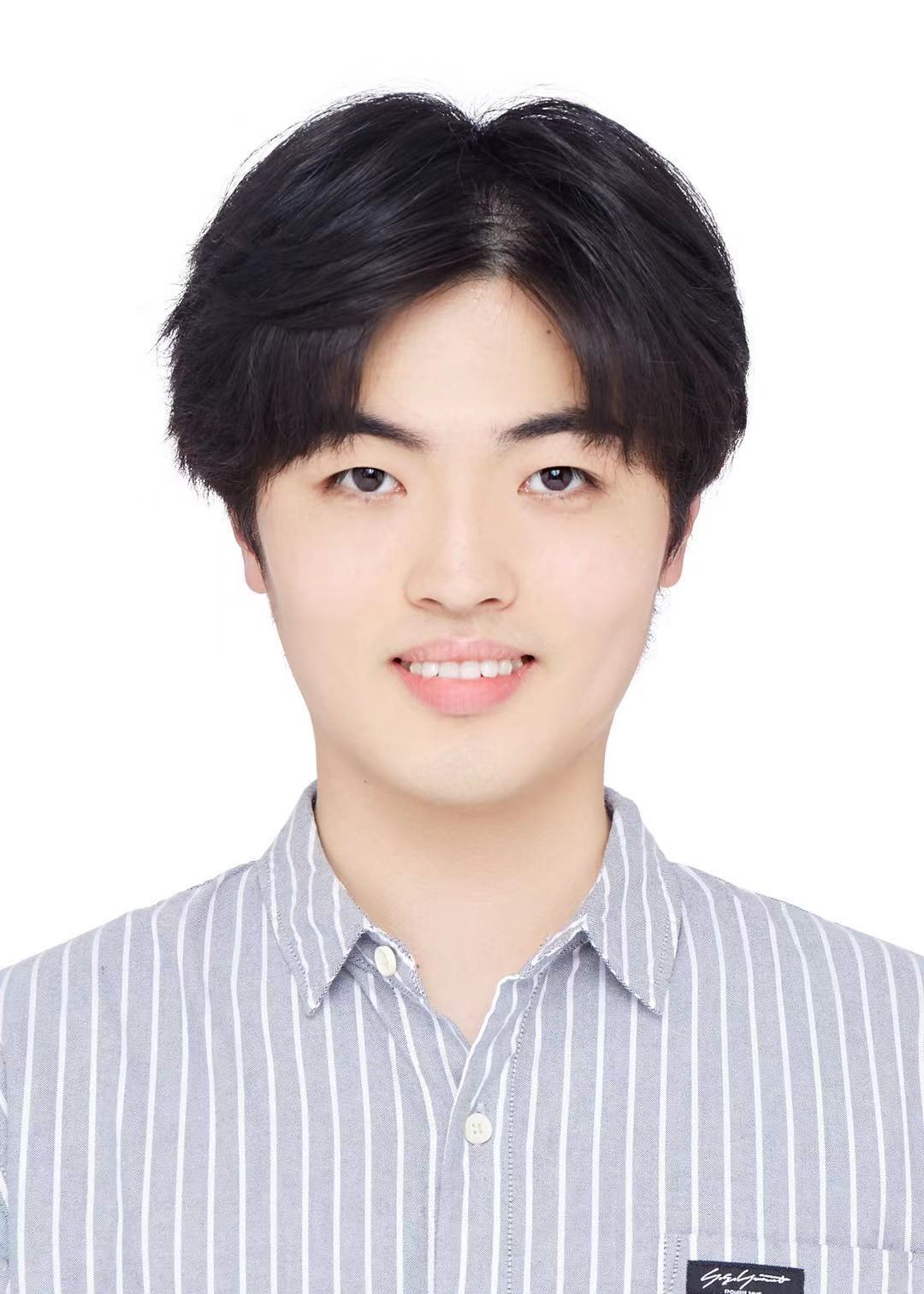}}]{Jinshu Chen}
received the B.S. degree in automation from Tongji University, Shanghai, China, in 2019. He is currently pursuing the M.S. degree in control science and engineering from Tongji University, Shanghai. His research interests include computer vision, deep learning and image processing.
\end{IEEEbiography}

\begin{IEEEbiography}[{\includegraphics[width=1in,height=1.25in,clip,keepaspectratio]{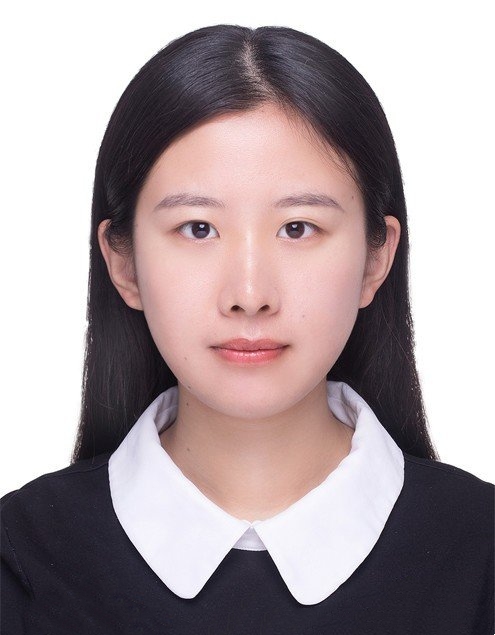}}]{Qihui Xu}
recevied the B.S.degree in automatic control from Tongji University, Shanghai, China, in 2020. She is currently pursuing the M.S. degree in control science and engineering from Tongji University, Shanghai. Her research focuses on the  use of deep learning technology for few-shot learning and image generation.
\end{IEEEbiography}

\begin{IEEEbiography}[{\includegraphics[width=1in,height=1.25in,clip,keepaspectratio]{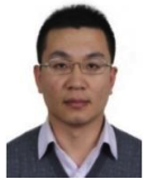}}]{Qi Kang}
(Senior Member, IEEE) received the B.S. degree in automatic control, the M.S. degree in control theory and control engineering, and the Ph.D. degree in control theory and control engineering from Tongji University, Shanghai, China, in 2002, 2005, and 2009, respectively.
From 2007 to 2008, he was a Research Associate with the University of Illinois, Chicago, IL, USA. From 2014 to 2015, he was a Visiting Scholar with the New Jersey Institute of Technology, Newark, NJ, USA. He is currently a Professor with the Department of Control Science and Engineering and the Shanghai Institute of Intelligent Science and Technology, Tongji University, Shanghai, China. His interests are in swarm intelligence, evolutionary computation, machine learning, and intelligent control and optimization in transportation, energy, and water systems.
\end{IEEEbiography}

\begin{IEEEbiography}[{\includegraphics[width=1in,height=1.25in,clip,keepaspectratio]{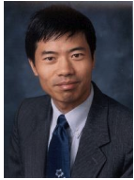}}]{MengChu Zhou}
(Fellow, IEEE) joined the New Jersey Institute of Technology (NJIT), Newark, NJ, USA, in 1990, where he is currently a Distinguished Professor. He has over 900 publications, including 12 books, over 600 journal articles (over 500 in IEEE TRANSACTIONS), 27 patents, and 29 book chapters. His interests are in Petri nets, intelligent automation, the Internet of Things, and big data. Prof. Zhou is fellow of International Federation of Automatic Control, American Association for the Advancement of Science and Chinese Association of Automation. He was a recipient of the Humboldt Research Award for U.S. Senior Scientists from the Alexander von Humboldt Foundation, the Franklin V. Taylor Memorial Award and the Norbert Wiener Award from the IEEE Systems, Man and Cybernetics Society, the Excellence in Research Prize and Medal from NJIT, and the Edison Patent Award from the Research and Development Council of New Jersey. He is also the Founding Editor of the IEEE Press Book Series on Systems Science and Engineering and the Editor-in-Chief of IEEE/CAA JOURNAL OF AUTOMATICA SINICA.
\end{IEEEbiography}




\end{document}